\documentclass{article}

\usepackage{times}
\usepackage{graphicx} 

\usepackage{natbib}

\usepackage{psfrag}
\usepackage{tikz}
\usepackage{adjustbox}
\usepackage{hyperref}
\usepackage{amsmath}
\usepackage{amssymb}
\usepackage{abstract}

\usepackage[accepted]{icml2015}

\let\footnotesize\small 

\icmltitlerunning{Max-Margin Rank-Likelihood}

\def\A{{\bf A}}
\def\a{{\bf a}}

\def\h{{\bf h}}
\def\I{{\bf I}}

\def\W{{\bf W}}
\def\w{{\bf w}}
\def\X{{\bf X}}
\def\x{{\bf x}}

\def\y{{\bf y}}
\def\Z{{\bf Z}}
\def\z{{\bf z}}

\def\0{{\bf 0}}
\def\1{{\bf 1}}

\def\Re{\mathbb{R}}

\def\<{\, \langle \,}
\def\>{\, \rangle \,}
\def\inv{^{-1}}
\def\ts{^\top}

\def\thalf{\tfrac{1}{2}}

\def\sign{\mathrm{sign}}
\def\max{\mathrm{max}}
\def\min{\mathrm{min}}

\def\bbeta{\boldsymbol{\beta}}

\def\bmu{\boldsymbol{\mu}}

\def\btheta{\boldsymbol{\theta}}

\def\DN{\mathcal{N}}
\def\TN{\mathcal{TN}}

\def\Ga{\mathrm{Gamma}}

\newcommand{\bs}{\backslash}

\newcommand{\ie}{i.e.,\ }
\newcommand{\eg}{e.g.,\ }
\newcommand{\wrt}{w.r.t.\ }

\newcommand{\quotes}[1]{``#1''}

\def\ctilde{\kern -.04em\lower .7ex\hbox{\~{}}\kern .04em}

\def\Ga{\mathrm{Ga}}
\def\rank{\mathrm{rank}}
\begin{document} 

\twocolumn[
\icmltitle{Non-Gaussian Discriminative Factor Models \\ via the Max-Margin Rank-Likelihood}


\vspace{-3mm}
\icmlauthor{Xin Yuan$^*$}
{eiexyuan@gmail.com}
\vspace{-0.5mm}
\icmlauthor{Ricardo Henao$^*$}
{r.henao@duke.edu}
\vspace{-0.5mm}
\icmlauthor{Ephraim L. Tsalik}
{e.t@duke.edu}
\vspace{-0.5mm}
\icmladdress{Duke University, Durham, NC, 27708, USA}
\vspace{-3mm}
\icmlauthor{Raymond J. Langley}
{rlangley@lrri.org}
\vspace{-0.5mm}
\icmladdress{Department of Immunology, Lovelace Respiratory Research Institute, Albuquerque, NM 87108, USA}
\vspace{-3mm}
\icmlauthor{Lawrence Carin}
{lcarin@duke.edu}
\vspace{-0.5mm}
\icmladdress{Duke University, Durham, NC, 27708, USA}

\icmlkeywords{Non-Gaussian, Max-Margin, Bayesian, Rank-likelihood}

\vskip 0.3in
]

{
	\renewcommand{\thefootnote}%
	{\fnsymbol{footnote}}
	\footnotetext[1]{Equal contribution.}
}
\begin{abstract}
	We consider the problem of discriminative factor analysis for data that are in general {\em non-Gaussian}. A Bayesian model based on the {\em ranks} of the data is proposed. We first introduce a new {\em max-margin} version of the rank-likelihood. A discriminative factor model is then developed, integrating the max-margin rank-likelihood and (linear) Bayesian support vector machines, which are also built on the max-margin principle. The discriminative factor model is further extended to the {\em nonlinear} case through mixtures of local linear classifiers, via Dirichlet processes. Fully local conjugacy of the model yields efficient inference with both Markov Chain Monte Carlo and variational Bayes approaches. Extensive experiments on benchmark and real data demonstrate superior performance of the proposed model and its potential for applications in computational biology.
\end{abstract}

\section{Introduction}
Modern applications in computational biology and bioinformatics routinely involve data coming from different sources, measured and quantified in different ways, e.g., averaged intensities in gene expression, cytokines and proteomics, or fragment counts in RNA and microRNA sequencing. However, they all share a common trait: data are rarely Gaussian, and they are often discrete, the latter due to digital technologies used for quantification. Nevertheless, a large proportion of statistical analyses performed on these data assume Gaussianity in one way or another. This is because customary preprocessing pipelines employ normalization and/or domain transformation approaches aimed at making the data as Gaussian, or at least as symmetric, as possible. For example, one popular yet simple strategy for RNA sequencing data is to rescale each sample to correct for technical variability, followed by log-transformation or quantile normalization \citep{dillies13a}. This and many other examples have the same rationale: the data transformations are {\em order preserving}, while also trying to achieve a desired distribution, typically Gaussian. 
\begin{figure}[htbp!]
	\centering
	\begin{psfrags}
		\psfrag{ds}[c][c][0.6][0]{density}
		\psfrag{rx}[c][c][0.6][0]{$\rank(\x)$}
		\psfrag{lx}[c][c][0.6][0]{$\log(\x$)}
		\psfrag{activeeTB}[c][c][0.55][0]{\hspace{-6pt}Active TB}
		\psfrag{latenttTB}[c][c][0.55][0]{\hspace{-2pt}Latent TB}
		\includegraphics[scale = 0.33]{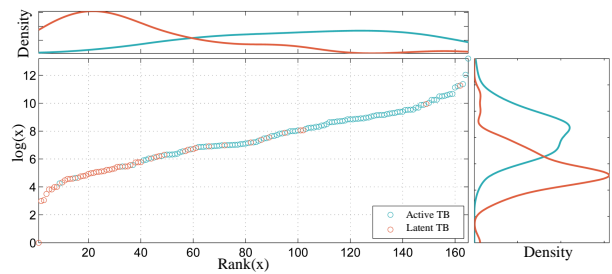}
	\end{psfrags}
	\vspace{-3mm}
	\caption{Intuition behind data modeling with ranks. Top and right panels are group-wise empirical distributions for ${\rank}(\x)$ and $\log(\x)$, respectively.}\label{fg:ranks}
	\vspace{-2mm}
\end{figure}
Figure~\ref{fg:ranks} shows expression for a particular gene and two phenotypes (active and latent tuberculosis), from the dataset described in Section~\ref{sc:tb}. The horizontal and vertical axes show respectively $\log(\x)$ (log-transformed) and $\rank(\x)$ (ranked) gene intensities. We see that from either axis we could derive a decision rule to separate the two groups, so that we can predict the label of new data, without worrying too much about the actual values or scaling of the axes. In fact, from their group-wise empirical distributions, we see that $\log(\x)$ and $\rank(\x)$ seem to have nearly the same optimal decision rule. Note that in general we are not required to log-transform the data, and in principle we could either use raw data or any other monotone transformation of the gene intensities while still being able to build a classifier, if the data support it. Motivated by this fact, and also by the success of standard nonparametric statistical approaches based on ranks, such as Spearman's rank correlation and Wilcoxon rank-sum test \citep{lehmann06a}, in this paper, we propose a new {\em discriminative factor model} based directly on the {\em ranks} of observed data as opposed to their values. 
This new model enjoys three significant benefits:
\begin{list}{\labelitemi}{\leftmargin=16pt \topsep=0pt \parsep=0pt}
	\item[(1)] Since we do not model actual values, we could treat ordinal, continuous and discrete data within the same framework. 
	\item[(2)] We can in principle make weaker assumptions about the distribution of the data.
	\item[(3)] We can jointly identify subsets of variables with similar (rank) correlation structures, some of which might be able to (partially) separate different classes and could be combined to build a classification model.
\end{list} 
These advantages come with the price of not being able to account for the actual values of the data, which is not such a big disadvantage, as long as we are only interested in parameters of the model involving {\em relative} differences or similarities between elements of a given dataset (which is typically the case when building classifiers).

Modeling with ranks is not a new idea, in fact \citet{Pettitt82rank} presented a linear regression model using a likelihood function based on the ranks of observed data, coined by the authors as {\em rank-likelihood}. More recently it was also used by \citet{hoff07a} to estimate the correlation matrix of data from disparate types, \eg binary, discrete and continuous. Here we employ the rank-likelihood as a building block for a discriminative factor model, with the ultimate goal of being able to jointly perform feature extraction and classification while decreasing the effort required to preprocess raw data.

The contributions of this paper are three-fold:
\begin{list}{\labelitemi}{\leftmargin=16pt \topsep=0pt \parsep=0pt}
	\item[(1)] We introduce a new {\em max-margin} version of the rank-likelihood geared towards Bayesian factor modeling, and we present a novel data augmentation scheme that allows for fast inference due to local conjugacy. 
	\item[(2)] We propose a discriminative factor model by integrating max-margin rank-likelihood, (linear) Bayesian support vector machines (SVMs) and global-local shrinkage priors. One key feature of our model is that likelihood functions for both data and labels have the max-margin property. 
	\item[(3)] We extend the discriminative factor model to {\em nonlinear} decision functions, through a mixture of local linear classifiers implemented via a Dirichlet process imposed on the latent space of the factor model.
\end{list} 

Experiments on benchmark and real data, namely USPS, MNIST, gene expression and RNA sequencing, demonstrate that the proposed model often performs better than competing approaches. Results on the real data demonstrate the potential of our model for applications in computational biology, not only for well-established, high-throughput technologies such as gene expression and metabolomics, but also in emerging ones such as RNA sequencing and proteomics.

\section{Max-margin rank-likelihood}
\paragraph{Ordinal probit model} Consider $N$ data samples, each a $d$-dimensional vector with ordinal values; the discussion of ordinal data helps to motivate and explain the model, which is subsequently generalized to real-valued data. The data are represented by the $d\times N$ matrix $\X$, the $n$th column of which represents the $n$th data vector. Let $x_{i,n}$ represent element $(i,n)$ of $\X$, corresponding to component $i$ of the $n$th data vector, modeled as
\begin{align}\label{eq:gfa}
x_{i,n} &= \ g_i(w_{i,n}) \,, \\
w_{i,n} &= \ \a_i\ts\z_n + v_{i,n} \,, \quad v_{i,n} \sim \ \DN(0,1) \,, \nonumber
\end{align}
where $\A=[\a_1 \ \dots \ \a_d]\ts\in \mathbb{R}^{d\times K}$ is the factor loadings matrix with $K$ factors, the factor scores for all $N$ data samples are represented by $\Z=[\z_1 \ \dots \ \z_N ]\in\mathbb{R}^{K\times N}$, $w_{i,n}$ is element $(i,n)$ of $\W\in\mathbb{R}^{d\times N}$, and $g_i(\cdot)$ is a non-decreasing function (such that the rankings of the $N$ realizations of component $i$ are preserved). Specifically, large values in $\x_i$ (rows of $\X$) correspond to large values in $\w_i$ (rows of $\W$).

Assume that component $i$ of each data vector takes values in the set $\{1,\dots, J_i\}$. Then function $g_i(\cdot)$ can be fully specified by $J_i-1$ ordered parameters $h_{i,1}<\cdots<h_{i,J_i-1}$, often called \quotes{cut points}, yielding
\begin{align}\label{eq:oprobit}
x_{i,n} = g_i(w_{i,n}) = j \quad {\rm if} \quad h_{i,j-1} < w_{i,n} < h_{i,j} \,,
\end{align}
where $h_{i,0}=-\infty$, $h_{i,J_i}=\infty$ and $\h_i=[h_{i,1} \ \ldots \ h_{i,J_i-1}]$ is a vector of thresholds for the $i$th row of $\W$. Equations~\eqref{eq:gfa} and \eqref{eq:oprobit} define a typical probit factor model for ordinal vector data \citep{hoff2009}.

Inferring $\W$ for the model in~\eqref{eq:gfa} and \eqref{eq:oprobit} is not complicated, because its conditional posterior corresponds to a truncated Gaussian distributions, \ie $w_{i,n}\sim \TN(\a_i\ts\z_n,1,h_{i,j-1},h_{i,j})$, where $h_{i,j-1}<w_{i,n}<h_{i,j}$ for $j=x_{i,n}$. Nevertheless, the model has three important shortcomings: ($i$) Specifying a prior distribution for $\{\h_i\}_{i=1}^d$ might be difficult because often such information is not available to the practitioner; ($ii$) if $J_i$  is large, the number of parameters of the model that need to be estimated increases substantially, making prior specification and inference harder; ($iii$) sampling from a truncated Gaussian distribution can be relatively expensive, and may be prone to numerical instabilities, especially when samples lie near the tails of the distribution.

\paragraph{Rank-likelihood model} Provided that $g_i(w_{i,n})$ is non-decreasing by assumption, we know that given $x_{i,n}<x_{i,n'}$, then $g_i(w_{i,n})<g_i(w_{i,n'})$ and $w_{i,n}<w_{i,n'}$, thus
\begin{align}\label{eq:ranklik}
\resizebox{.90\hsize}{!}{$\displaystyle
	R(\x_i)=\{\w_i\in\Re^N:w_{i,n}<w_{i,n'} \ {\rm if} \ x_{i,n}<x_{i,n'} \} \,,
	$}
\end{align}
where $R(\x_i)$ is the set of all possible vectors $\w_i$ such that $\rank(\x_i)=\rank(\w_i)$. Given $\x_i$, since neither $\w_i$ nor $p(\w_i\in R(\x_i))$ depend on $g_i(\w_i)$, we can formulate inference for $\A$ and $\Z$ directly in terms of $\w_i\in R(\x_i)$ through the {\em rank-likelihood} representation $p(\w_i\in R(\x_i)|\A,\Z)$ \citep{Pettitt82rank}. Specifically, we can write the joint probability distribution of the model above as
\begin{adjustbox}{minipage=1.05\linewidth,scale=0.95}
	\begin{align}\label{eq:joint}
	& \textstyle{\prod_{i=1}^d p(\w_i\in\ R(\x_i),\a_i,\Z) }= \nonumber \\
	& p(\A)p(\Z)\prod_{i=1}^d\left\{\int_{R(\x_i)}\prod_{n=1}^N \DN(w_{i,n}; \a_i\ts\z_n,1)d{w_{i,n}}\right\} \,.
	\end{align}
\end{adjustbox}

The integrals to the right hand side of \eqref{eq:joint} are in general difficult to compute. However, it is not necessary to do so, because we can estimate the posterior of parameters $\{\A,\Z,\W\}$ via Gibbs sampling, by iteratively cycling through their conditional posterior distributions. For $\A$ and $\Z$ we need to sample from $p(\Z|\W,\A)$ and $p(\A|\W,\Z)$, respectively, where we instantiate $\W$ such that $\w_i\in R(\x_i)$ for $i=1,\ldots,d$. For $\w_i$ we only need to be able to sample $\w_i$ from $p(\w_i\in R(\x_i)|\a_i,\Z)$. We can write $p([ w_{i,n} \ \w_{i\bs n} ]\in R(\x_i),\a_i,\Z)$, where $\w_{i\bs n}$ contains all elements from $\w_i$ but $w_{i,n}$. From \eqref{eq:gfa} and \eqref{eq:ranklik}, $w_{i,n}$ is Gaussian and restricted to the set $R(\x_i)$, respectively. Conditioning on $\z_{i\bs n}$, we have
\begin{align*}
p(w_{i,n}|\w_{i\bs n},\a_i,\z_n) & = \ p(w_{i,n}|w_{i,n}^l,w_{i,n}^u,\a_i,\z_n) \nonumber \\
& = \ \TN(\a_i\ts\z_n,1,w_{i,n}^l,w_{i,n}^u) \,,
\end{align*}
where 
%
%
$w_{i,n}^l = \max\{w_{i,n'}:x_{i,n'}<x_{i,n}\}$ and $w_{i,n}^u = \min\{w_{i,n'}:x_{i,n}<x_{i,n'}\}$,
which jointly guarantee that $[ w_{i,n} \ \w_{i\bs n} ]\in R(\x_i)$. Note that given $\{w_{i,n}^l,w_{i,n}^u\}$, $w_{i,n}$ is conditionally independent of $\w_i\bs\{w_{i,n}^l,w_{i,n}^u\}$ and also that the conditional posteriors for the ordered probit and rank-likelihood based models are identical except that for the former, constraints for $w_{i,n}$ come from $\w_{i\bs n}$ as opposed to thresholds $\h_i$. In fact, the rank-likelihood model can be seen as an alternative to the ordered probit model in which the threshold variables have been marginalized out \citep{hoff2009}. It is important to point out that in applications when the connection between observed and latent variables, $g_i(\w_i)$, is of interest, the rank-likelihood is not applicable. Fortunately, in factor models we are usually only interested in $\A$ and $\Z$, the loadings and the factor scores, respectively (see \citet{murray2013bayesian}, for example).
\paragraph{Max-margin rank-likelihood} 
\begin{figure}[t!]
	\centering
	\vspace{-1mm}
	\begin{tikzpicture}[scale=1.2]
	\tikzstyle{dp} = [circle,draw=black,fill=white,minimum size=1mm,inner sep=0pt]
	\tikzstyle{gr} = [dashed,thin,draw=black!50]
	
	\draw[->] (-0.2,0) -- (3.5,0) node[right] {$x$};
	\draw[->] (0,-0.2) -- (0,2.0) node[above] {$w$};
	
	\foreach \x/\xtext in { 0.5/x_{i,l}, 1.25/x_{i,n}, 2/x_{i,u}, 2.75/\ldots }
	\draw[shift={(\x,0)}] (0pt,2pt) -- (0pt,-2pt) node[below] {$\xtext$};
	
	\foreach \x/\y/\w in { 0.5/0.5/w_{i,n}^l, 1.25/1/w_{i,n}, 2/1.5/w_{i,n}^u } {
		\draw[gr] (0,\y) -- (\x,\y);
		\node[dp,label=right:$\w$] at (\x,\y) {};
	}
	
	\draw[thin,red] (1.25,0.7) -- (1.25,1.3) -- (1.75,1.8);
	\draw[thin,red] (1.25,0.7) -- (1.75,0.2);
	\draw[->,red] (1.25,0) -- (1.25,2) -- (2.25,2) node[right] {$\ell_\epsilon^u+\ell_\epsilon^l$ (loss function)};
	
	\draw[gr] (0,1.3) -- (1.25,1.3); \node[left] at (0,1.4) {$\epsilon$};
	\draw[gr] (0,0.7) -- (1.25,0.7); \node[left] at (0,0.6) {$\epsilon$};
	\end{tikzpicture}
	\vspace{-4mm}
	\caption{Graphical representation of the loss function associated to the max-margin rank-likelihood in~\eqref{eq:plik}, where $\ell_\epsilon^u+\ell_\epsilon^l=\ell_\epsilon(w_{i,n}-w_{i,n}^u)+\ell_\epsilon(w_{i,n}^l-w_{i,n})$. Note that $w_{i,n}^l+\epsilon<w_{i,n}<w_{i,n}^u-\epsilon$ is not penalized by the loss function.}\label{fg:loss}
	\vspace{-4mm}
\end{figure}
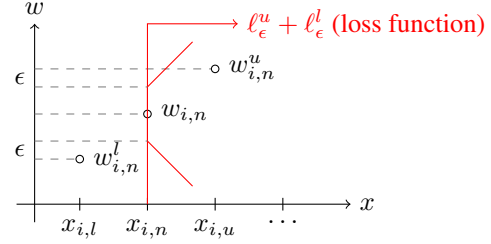
One disadvantage of the rank-likelihood model is that differences between elements of $\w_i$ can be {\em arbitrarily small}, as there is no mechanism in the prior distribution of $\w_i$ to prevent this from happening. In the ordered probit model we can do so via the prior for the thresholds $\h_i$, however this is not necessarily easily accomplished. Fortunately, for the rank-likelihood we can alleviate this issue by modifying \eqref{eq:ranklik} as
\begin{align}\label{eq:mmranklik}
\resizebox{.90\hsize}{!}{$\displaystyle
	R_{\rm mm}(\x_i)=\{\w_i\in\Re^N:w_{i,n}<w_{i,n'}-\epsilon \ {\rm if} \ x_{i,n}<x_{i,n'} \} \,,
	$}
\end{align}
where we have made explicit that any two distinct elements of $\w_i$ must be separated by a gap of size no smaller than $\epsilon>0$. Furthermore, $\max\{w_{i,n'}:x_{i,n'}<x_{i,n}\}+\epsilon<w_{i,n}<\min\{w_{i,n'}:x_{i,n}<x_{i,n'}\}-\epsilon$. From \eqref{eq:mmranklik} we can write a pseudo-likelihood for $w_{i,n}$ as 
\begin{align}\label{eq:plik}
\resizebox{.90\hsize}{!}{$\displaystyle
	L_i(w_{i,n}|\w_{i\bs n}) = \exp\left\{-\ell_\epsilon(w_{i,n}-w_{i,n}^u)-\ell_\epsilon(w_{i,n}^l-w_{i,n})\right\} \,,
	$}
\end{align}
where $w_{i,n}=\a_i\ts\z_n$, $w_{i,n}^u=\a_i\ts\z_n^u$, $w_{i,n}^l=\a_i\ts\z_n^l$ and $\ell_\epsilon(u)=2\max(0,u+\epsilon)$ can be interpreted as the \quotes{one-sided} $\epsilon$-sensitive loss. From \eqref{eq:plik} this means that $\ell_\epsilon(u)>0$ only if $w_{i,n}<w_{i,n}^l+\epsilon$ or $w_{i,n}>w_{i,n}^u-\epsilon$; it also means that this loss function does not penalize $w_{i,n}^l+\epsilon<w_{i,n}<w_{i,n}^u-\epsilon$ and $\epsilon$ is called the {\em margin}. See Figure~\ref{fg:loss} for a graphical representation of the proposed composite loss function. Maximizing \eqref{eq:plik} is equivalent to finding $\w_i\in R_{\rm mm}(\x_i)$ such that differences between neighbor elements of $\w_i$ are maximal given $\epsilon$, hence the term {\em max-margin} is used.

Recent work by \citet{polson11a} has shown that $\ell_\epsilon(u)$ admits a location-scale mixture of Gaussians, specifically, they showed that $\exp\{-2\max(0,u)\}=\int\DN(u;-\lambda,\lambda)d\lambda$, thus we can rewrite \eqref{eq:plik} as
\begin{align}\label{eq:plik_gauss}
& L_i(w_{i,n}|\w_{i\bs n})= \int \DN(w_{i,n}-w_{i,n}^u;-\epsilon-\lambda_{i,n}^u,\lambda_{i,n}^u) \nonumber \\
&\qquad\times \DN(w_{i,n}^l-w_{i,n};-\epsilon-\lambda_{i,n}^l,\lambda_{i,n}^l)d\lambda_{i,n}^u d\lambda_{i,n}^l \,,
\end{align}
where $\DN(u;\cdot)$ is the density function of a Gaussian distribution, and we have introduced two sets of latent variables $\{\lambda_{i,n}^u\}$ and $\{\lambda_{i,n}^l\}$. This {\em data augmentation} scheme implies that the pseudo-likelihood before marginalization, $L_i(w_{i,n}|\w_{i\bs n},\lambda_{i,n}^u,\lambda_{i,n}^l)$, is conjugate to a Gaussian distribution, just as in the original rank-likelihood formulation, but without the difficulty of truncated Gaussians, because $w_{i,n}$ is now exactly $\a_i\ts\z_n$, not a random variable. Note that as a result of this, we have transferred the {\em uncertainty} between the rank of $x_{i,n}$ and the factorization $\a_i\ts\z_n$ from $w_{i,n}$ in the rank-likelihood (and the ordered probit) to the set of location-scale parameters $\{\lambda_{i,n}^u,\lambda_{i,n}^l\}$ in our max-margin formulation.
\paragraph{Discrete and continuous data} So far we have assumed that we have ordinal data (the cut points of the ordinal model help explain the associated mechanics of the rank-likelihood model). However, we can also use the rank-likelihood with discrete or continuous data, as long as we acknowledge that likelihood and posteriors derived from them only contain information about the order of the observations and not their actual values.

In general terms, factor models are concerned with learning about the covariance structure of observed data via a low rank matrix decomposition, $\A\Z$. In this sense, the role of the likelihood is to define the way in which covariances are measured. This means that one important difference between standard Gaussian and rank-likelihood based factor models is that they use different notions of covariance; very much in the same spirit of the differences between Pearson and Spearman correlations. Another important difference is the generative mechanism implied by the likelihood. In the rank-likelihood, we can generate a statistic, namely the rank, based on a sample population but not their values. This happens because we ignore the part of the model that links the statistic with actual data, specifically
\begin{align*}
& p(\x_i|\a_i,\Z,\h_i) = p(\rank(\x_i),\x_i|\a_i,\Z,\h_i)\nonumber \\
& \qquad \qquad= p(\rank(\x_i)|\a_i,\Z)p(\x_i|\rank(\x_i),\a_i,\Z,\h_i) \,.
\end{align*}
We infer $\a_i$ and $\Z$ strictly from $p(\rank(\x_i)|\a_i,\Z)$, the marginal likelihood, via $\w_i\in R_{\rm mm}(\x_i)$. Since we ignore $p(\x_i|\rank(\x_i),\a_i,\Z,\h_i)$, we do not infer the thresholds $\h_i$, thus in the strictest sense we are not using all information provided by $\x_i$, \ie its values, however we are assuming that ranks alone contain enough information to be able to characterize the covariance structure of the data so we can reliably estimate $\A$ and $\Z$. Additional examples and further discussion of Bayesian analysis employing similar marginal likelihood strategies can be found in \citet{monahan92a}.

\section{Bayesian SVM based discriminative factor model}
When the data being analyzed belong to two different classes, encoded as $\{-1,1\}$, labels $\y = [y_1 \ \dots \ y_N]\ts \in \{-1,1\}$ will encourage our factor model to learn discriminative features (loadings and scores) from the data, then these features can be used to make predictions for new data. This modeling approach is commonly known as supervised dictionary learning or discriminative factor analysis \citep{Mairal08superviseddictionary}. From a Bayesian perspective, factor models and probit/logit link based classifiers have been already successfully combined; see for instance \citet{Salazar12ICML,Quadriantos13IBP}.

Unlike previous work, we continue with the max-margin theme and develop a supervised factor model using Bayesian support vector machines (SVMs). The same result from \citet{polson11a} used above to derive the max-margin rank-likelihood provides a pseudo-likelihood for the hinge loss, traditionally employed in the context of SVMs~\citep{polson11a}. Specifically,
\begin{align}\label{eq:hingelik}
& L_n(y_n|\bbeta,\z_n) = \exp\{-2 \max (0,u_n)\} \nonumber\\
& \quad= \int_0^{\infty} \frac{1}{\sqrt{2\pi \lambda^c_n}} \exp\left( -\frac{1}{2} \frac{(u_n + \lambda^c_n)^2}{\lambda^c_n}\right) d\lambda^c_n \,,
\end{align}
where $u_n=1-y_n\bbeta\ts\z_n$, $\bbeta\in\Re^K$ is a vector of classifier coefficients and $\{\lambda_n^c\}$ is a vector of latent variables, with superscript $c$ denoting the classifier. In \citet{polson11a} covariates, $\z_n$, are observed while here they are latent variables (factor scores) that need to be estimated jointly with the remaining parameters of a factor model. It has been shown empirically that linear margin-based classifiers, SVMs being a special case, often perform better than those using logit or probit links \citep{polson11Rejoinder,Henao2014}.

Interestingly, in our factor model the max-margin mechanism plays two roles, \ie data and labels are both connected to the factor model core via max-margin pseudo-likelihoods: rank-likelihood for the data and hinge loss for the labels. Furthermore, for the loadings, since shrinkage for $\A$ is usually a requirement for interpretability or when $N\ll d$, here we use a three-parameter-beta normal prior (${\cal TPBN}$)~\citep{Armagan12}, a fairly general global-local shrinkage prior~\citep{polson10a}, for which it has been demonstrated that it has better mixing properties than priors such as spike-slab~\citep{Carvalho_horseshoe}. Shrinkage for the elements of $\bbeta$ is also employed, because it allows us to identify the features of $\A$ that contribute to the classification task. Intuitively, we can see $\A$ as a dictionary with $K$ features, each feature explaining a subset of the input variables due to shrinkage; via separate shrinkage within the model, $\bbeta$ selects from the $K$ features to build a predictor for labels $\y$. Being able to specify global and local properties independently makes the ${\cal TPBN}$ prior attractive for high-dimensional settings, such as gene expression and RNA sequencing, which are precisely the types of data we will focus on in our experiments.
\paragraph{Linear discriminative factor model} By imposing the max-margin rank-likelihood construction in \eqref{eq:mmranklik} to $\X$ and the hinge loss to $\y$ via pseudo-likelihoods in \eqref{eq:plik_gauss} and \eqref{eq:hingelik}, respectively, one possible prior specification for the supervised factor model parameterized by $\{\A,\Z,\bbeta\}$ is
\begin{align*}
a_{i,k} \ & \sim \ {\cal TPBN}(r_a,s_a, \Phi^{(a)}_k) \,, \quad \z_n \ \sim \ \DN(0, \I_K) \,, \nonumber\\
\beta_{k} \ & \sim \ {\cal TPBN}(r_{\beta},s_{\beta},\Phi^{(\beta)}) \,, 
\end{align*}
where $\Phi^{(a)}_k, \Phi^{(\beta)}$ are global shrinkage parameters for loadings $\A$ and classifier coefficients $\bbeta$. Furthermore, for the ${\cal TPBN}$ prior we can write
\begin{align*}
a_{i,k}  \ \sim & \ \DN(0, \xi_{i,k}) \,,  & \quad \beta_{k} \ \sim & \ \DN(0, b_k) \,, \\
\xi_{i,k} \ \sim & \ \Ga(r_a, \eta_{i,k}) \,,  & \quad b_k \ \sim & \ \Ga(r_{\beta}, e_k) \,, \\
\eta_{i,k} \ \sim & \ \Ga(s_a,\Phi^{(a)}_k ) \,, & \quad e_k \ \sim & \ \Ga(s_{\beta},\Phi^{(\beta)}) \,.
\end{align*}
Setting $r_a=s_a=\thalf$ (for $\bbeta$, it is $r_{\beta}=s_{\beta}=\thalf$), a special case of ${\cal TPBN}$ corresponds to the widely known horseshoe prior~\citep{Carvalho_horseshoe}. Note that each column of the loadings has a different global shrinkage parameter $\textstyle \Phi^{(a)}_k$, thus allowing them to have different degrees of shrinkage. We can also infer $\Phi^{(a)}_k$ (and $\Phi^{(\beta)}$) by letting $\Phi^{(a)}_k \sim \Ga(\thalf,\tilde{\Phi})$ and $\tilde{\Phi} \sim {\rm Ga}(\thalf,1)$. As a result of having individual shrinkage parameters for each column of $\A$, we could say that the prior is capable of \quotes{turning off} unnecessary factors, hence having an automatic relevance determination flavor to it \citep{mackay95a,Wipf08anew}. This is indeed the behavior we see in practice; there are other ways to select the number of factors, e.g., by adding a multiplicative gamma prior to matrix $\A$~\citep{Dunson11MultiGam}.

\vspace{-3mm}
\paragraph{Non-Linear discriminative factor model} When the latent space for factor scores, $\z_n$, is not linearly separable, nonlinear classification approaches might be more appropriate. One may use a kernel to extend the linear SVM to its nonlinear counterpart. However, from a Bayesian factor modeling perspective, adding kernel-based nonlinear classifiers is nontrivial, because they tend to make inference complicated and computationally expensive due to loss of conjugacy for the parameters involved in the nonlinear component of the model, \ie the kernel function. From a different perspective, it is still possible to build a {\em global nonlinear} decision rule as a mixture of {\em local linear} classifiers \citep{shahbaba09a,Fu10TNN}. The basic idea is to assume factor scores as coming from a mixture model, in which each mixture component has an associated {\em local} linear Bayesian SVM. Here we use a Dirichlet process (DP) in its stick-breaking construction~\citep{Sethuraman01DP}, represented as
\begin{align} \label{eq:DP}
G & = \textstyle{\sum_{t=1}^{\infty} q_t \delta_{\btheta^*_t}} \,, 
& \quad q_t & = \textstyle{\nu_t\prod_{l=1}^{t-1}(1-\nu_l) }\,, \nonumber \\
\nu_t & \sim {\rm Beta}(1,\alpha) \,, & \quad \btheta_t^* & \sim G_0 \,,
\end{align}
where $\sum_{t=1}^{\infty}q_t = 1$, $\delta_{\btheta^*_t}$ represents a point measure at $\btheta_t^*$ and $\alpha$ is the concentration parameter. Applied to our model, factor scores and labels are drawn from a parametric model $y_n,\z_n\sim f(\btheta_n)$ with parameters $\btheta_n$, where $\btheta_n\sim G$. For $G$ as in \eqref{eq:DP} and a finite number of samples $N$, many of the $\{y_n,\z_n\}$ share the same parameters, therefore making $\{y_n,\z_n\}$ a draw from a mixture model. Specifically, we make 
$f(\btheta_n)=L_n(y_n|\bbeta_n,\z_n)\DN(\z_n|\bmu_n,\psi_n\inv\I_K)$, $G_0={\cal TPBN}(\bbeta|r_\beta,s_\beta,\Phi^{(\beta)})\times\DN(\bmu|\0,\I_K)\times\Ga(\psi|\psi_s,\psi_r)$ 
%
and $\{\bbeta_n,\bmu_n,\psi_n\}=\{\bbeta_t,\bmu_t,\psi_t\}$ if sample $n$ belongs to the $t$-th component of the mixture. In practice we truncate the sum in~\eqref{eq:DP} to $T$ terms to make inference easier \citep{Ishwaran01DP} and set $\psi_s=1.1$ and $\psi_r=0.001$ (\ie a non-informative prior).

\vspace{-3mm}
\paragraph{Predictions} Making predictions for new data using our model is conceptually simple. We use the pair $\{\y,\X\}$ to estimate the parameters of the model (training), namely $\{\A,\Z,\bbeta\}$, then given a test point $\x_\star$, we go through three steps: ($i$) Compare $\x_\star$ to $\X$ to determine the rank of each component of $\x_\star$ \wrt to the training data, which amounts to finding $\{w_{i,\star}^l,w_{i,\star}^u\}$, for $i=1,\ldots,d$. ($ii$) For fixed $\{\A,w_{i,\star}^l,w_{i,\star}^u\}$, estimate $\z_\star$ from its conditional posterior. ($iii$) Make a prediction for $\x_\star$ using $\sign(\bbeta\ts\z_\star)$. 

The first step of this prediction process is exclusive to the proposed rank-likelihood model, and implies that we are required to keep the training data in order to make predictions. This is in the same spirit of supervised kernel methods, in the sense that predictions are a function of the data used to fit the model \citep{scholkopf01a}. Note, however, that for a single component of a test point, $x_{i,\star}$, we only need two elements of the training set: the two elements from $\x_i$ closest to $x_{i,\star}$ from above and below, which is closely related to the $k$-nearest neighbor paradigm (rather than $k$ nearest neighbors, we only require the two training neighbors ``to the left and right'' of a test data component). Intuitively, what our model does at prediction is to find a latent representation $\z_\star$ such that $\x_\star$ is in between but as far as possible from its upper and lower bounds \wrt to $\X$. This is a very unique characteristic of our model.

\vspace{-3mm}
\paragraph{Inference} Due to fully local conjugacy, we can write the conditional posterior distribution for all parameters of our model in closed form, making Markov Chain Monte Carlo (MCMC) inference based on Gibbs sampling a straightforward procedure. 
Space limitations prevent us from presenting the complete set of conditionals, however below we show expressions for the parameters involving the max-margin rank-likelihood in \eqref{eq:plik_gauss}, namely $\A$ and $\Z$.
For convenience, we denote
\vspace{-3mm}
\begin{align*}
\Gamma_{k,n} & = \frac{y_n\beta_k [1+\lambda_n^c - y_n (\bbeta\ts\z_n)_{\bs k}] }{\lambda^c_n} \,, \\
\lambda_{i,n} & = (\lambda_{i,n}^l)\inv+(\lambda_{i,n}^u)\inv \,, \\ 
(\bbeta\ts\z_n)_{\bs k} & = \bbeta\ts\z_n-\beta_kz_{k,n} \,,
\end{align*}
In the following conditional posterior-distributions, \quotes{$\cdot$} refers to the conditioning parameters of the distributions.

Sampling ${\bf A}$:
\begin{align*}
p(a_{i,k}| \ \cdot)  &=  \ \DN(\mu_{a_{i,k}}, \sigma^2_{a_{i,k}}) \,,  \\
\sigma^{-2}_{a_{i,k}}& = \textstyle{ \xi_{i,k}^{-1} + \sum_{n=1}^N z^2_{k,n} \lambda_{i,n}\inv} \,,\\
\mu_{a_{i,k}} &=\textstyle{\sigma^2_{a_{i,k}} \sum_{n=1}^N z_{k,n} \Delta^{(k)}_{i,n}} \,,
\end{align*}
Sampling ${\bf Z}$:
\begin{align*}
p(z_{k,n}| \ \cdot) &=  \ \DN(\mu_{z_{k,n}}, \sigma^2_{z_{k,n}}) \,, \\
\sigma^{-2}_{z_{k,n}} &= \textstyle{1 + \sum_{i=1}^d  a^2_{i,k}\lambda_{i,n}\inv + \frac{\beta^2_k}{\lambda^c_n} }\,, \\
\mu_{z_{k,n}} &= \textstyle{\sigma^2_{z_{k,n}}\left(\sum_{i=1}^d a_{i,k} \Delta^{(k)}_{i,n} + \Gamma_{k,n}\right)} \,,
\end{align*}	
where
\begin{align*}
\Delta^{(k)}_{i,n} = \left(\tfrac{w_{i,n}^l + \epsilon - w_{i,n}}{\lambda^l_{i,n}} - \tfrac{w_{i,n} + \epsilon - w_{i,n}^u}{\lambda^u_{i,n}}\right) + a_{i,k} z_{k,n}\lambda_{i,n} \,.
\end{align*}

Conditional posteriors for the remaining parameters: $\{\lambda_{i,n}^l,\lambda_{i,n}^u,\lambda^c_n,\bbeta\}$ and $\{\xi_{i,k},\eta_{i,k},\Phi_k^{(a)},b_k,e_k,\Phi^{(\beta)}\}$ can be found in \citet{polson11a} and \citet{Armagan12}, respectively. 
In our experiments we set $\epsilon=0.05$, however a conjugate prior (gamma distribution) exists hence $\epsilon$ can be inferred if desired. Inference details for the DP specification for the factor scores can be found for instance in \citet{Ishwaran01DP,neal00a}.

In applications where speed is important, we can use all conditional posteriors including those above to derive a variational Bayes (VB) inference algorithm for our model, which loosely amounts to replacing the conditioning on variables with their corresponding moments. Details of the conditionals are not shown here for brevity, and details of the VB procedure are found in the Supplementary Material. 

\paragraph{Other related work} For ordinal data, \citet{xu2013fast} presented a factor model using the ordered probit mechanism, but in which the probit link is replaced with a max-margin pseudo-likelihood. Inference is very efficient, but they still have to infer the thresholds $\{\h_i\}$. However, in their collaborative prediction applications, variables only take one of six possible values. For count data, \citet{chib1998posterior} proposed a generalized-linear-model inspired Bayesian model for Poisson regression, that can be easily extended to a factor model. However, expensive Metropolis-Hastings sampling algorithms need to be used, due to the non-conjugacy between the prior for $\A$ and the log link. More recently, \citet{ZhouAISTATS12} presented a novel formulation of Poisson factor analysis (PFA), based on the beta-negative binomial process, for which inference is efficient. Furthermore, none of the approaches discussed above consider discriminative factors models, and for PFA this is very difficult, because in that case the prior for the factor scores is not a Gaussian distribution and is thus not conjugate to the SVM or probit-based likelihoods. As a result, building discriminative factor models under that framework is challenging, at least not without Metropolis-Hastings style inference.

\section{Experiments}
\begin{table}[tbp!]
	\caption{Composition of different methods.}
	\centering
	{\small
		\begin{tabular}{c|ccc}
			Method   & Likelihood & Classifier & DPM \\ 
			\hline G-L-Probit & Gaussian & probit & No \\
			G-L-BSVM & Gaussian & BSVM & No \\
			\hline OR-L-Probit & ordinary rank & probit & No \\
			OR-L-BSVM & ordinary rank & BSVM & No \\
			R-L-BSVM & max-margin rank & BSVM & No \\
			\hline G-NL-BSVM & Gaussian & BSVM & Yes \\
			R-NL-BSVM & max-margin rank & BSVM & Yes \\
		\end{tabular}}
		\label{Table:Method}
		\vspace{-3mm}
	\end{table}
	We present numerical results on two benchmark (USPS and MNIST) and two real (gene expression and RNA sequencing) datasets, using part of or all methods summarized in Table~\ref{Table:Method}; inference is performed via VB. The data {\em likelihood} can be either {\em Gaussian}, {\em rank} or the {\em max-margin} rank-likelihood. The labels (classifier) can be modeled using the {\em probit} link or the Bayesian SVM (BSVM) pseudo-likelihood. When the DP mixture (DPM) model is used, the classifier results in a nonlinear (NL) decision function. Everywhere we set $K=20$, $T=5$ and performance measures were averaged over 5 repetitions (standard deviations are also presented). We verified empirically that further increasing $K$ or $T$ does not significantly change the outcome of any of our models. All code used in the experiments was written in Matlab and executed on a 3.3GHz desktop with 16Gb RAM.
	
	\begin{table*}[t!]
		\caption{Mean error rates ($\%$) and runtime in seconds for the test data of the USPS 3 vs. 5 subtask.}
		\label{Table:Err_USPS}
		\centering
		\small{
			\begin{tabular}{c|cc|ccc|cc}
				& G-L-Probit & G-L-BSVM &  OR-L-Probit &  OR-L-BSVM & R-L-BSVM & G-NL-BSVM & R-NL-BSVM \\
				\hline
				Error & 5.95$\pm$0.005 & 5.86$\pm$0.008 &  5.05$\pm$0.013 &  4.92$\pm$0.027 & {\bf 4.53$\pm$0.026} & 3.88$\pm$0.017 & {\bf 3.23$\pm$0.025} \\
				Runtime &  8.64 & 10.29 & 14.07 & 14.19 & 16.05 &  23.81 & 36.63 \\
			\end{tabular}}
		\end{table*}
		\begin{table*}[t!]
			\caption{Mean error rates (\%) and runtime in seconds for the test data of the MNIST 3 vs. 5 subtask.}
			\label{Table:Err_MNIST}
			\centering
			{\small
				\begin{tabular}{c|ccc|ccc}
					& G-L-BSVM & R-L-BSVM & L-SVM & G-NL-BSVM & R-NL-BSVM & NL-SVM \\
					\hline
					Error  & 5.05$\pm$0.053  & 4.84$\pm$0.014 &  {\bf 4.68}& 4.21$\pm$0.010 & 2.10$\pm$0.007 & {\bf 2.00} \\
					Runtime & 150 & 220 &  140  & 400 & 600 & 304 \\
				\end{tabular}}
			\end{table*}
			In the following experiments we focus on comparing discriminative factor models against each other to show how each component of the model contributes to the end performance produced by our full model. In particular, we show that the Bayesian SVM, max-margin rank-likelihood and nonlinear decision function all improve the overall performance on their own, when compared to standard approaches such as probit regression and Gaussian models on log-transformed data.
			It is important to take into consideration that our model is at the same time trying to explain the data and to build a classifier via a linear latent representation of ranks, thus we will not attempt to match results obtained by more sophisticated state-of-the-art classification models (e.g., a nonlinear SVM applied directly to raw data may yield a good classifier, but it does not afford the generative interpretability of a factor model, the latter particularly relevant to biological applications). Our model is ultimately trying to find a good {\em balance} between covariance structure modeling, interpretability through shrinkage and classification performance. All of this is done with the very important additional benefit of not requiring distributional assumptions about the values of the data, as this information is usually not known in practice (as in the subsequent biological experiments below). However, in those cases where the distribution is known \emph{a priori} it should certainly be reflected in likelihood function.
			%
			%
			%
			\subsection{Handwritten digits}
			Digitized images are a good example of essentially non-Gaussian data traditionally modeled using Gaussian noise models in the context of factor models and dictionary learning \citep{Mairal08superviseddictionary}. However, depending on preprocessing, they can be naturally treated either as continuous variables representing pixel intensities when filtering/smoothing is pre-applied, or as discrete variables representing pixel values when raw data is available. Our running hypothesis here is that a rank-likelihood representation for pixels is more expressive than its Gaussian counterpart. Intuitively, a discriminative factor model should be able to find features (subsets of representative pixels) that separate image subtypes. However, without the Gaussian assumption for observations, our rank model might be able to adapt to more general conditions, \eg skewed or heavy-tailed distributions. In our results we use classification error on a test set as a quantitative measure of performance.
			%
			\paragraph{USPS} First we consider the models in Table~\ref{Table:Method} to the well known 3 vs. 5 subtask of the USPS handwritten digits dataset. It consists of 1540 smoothed gray scale $16\times 16$ images rescaled to fit within the $[-1,1]$ interval. Each observation is a 256-dimensional vector of scaled pixel intensities. Here we use the resampled version, which is 767 images for model fitting and the remaining 773 for testing. Results in Table~\ref{Table:Err_USPS} show that consistently: rank-likelihood based models outperform Gaussian models, BSVM outperforms the probit link, and nonlinear outperforms linear classifiers. Furthermore, the proposed max-margin rank likelihood model performs best in both variants, linear and nonlinear. In every case inference took less than 1 minute.
			%
			\paragraph{MNIST} Next we consider the same 3 vs. 5 task, this time on a larger dataset, the MNIST database. The dataset is composed by 11552 training images and 1902 test images. Unlike USPS, MNIST consists of $28\times 28$ raw 8-bit encoded images, so each observation is a 784-dimensional vector of pixel values (discrete values from $\{0,\dots, 255\}$). Results for four out of six methods from Table~\ref{Table:Method} are summarized in Table~\ref{Table:Err_MNIST}. Results for probit based models were not as good as those for BSVM, thus not showed here, nor in the upcoming experiments. Instead, we include results for a linear (L-SVM) and nonlinear (NL-SVM) SVM with RBF kernel directly applied to the data as baselines, without the factor model. From Table~\ref{Table:Err_MNIST} we see that the proposed model works better than the Gaussian model, and that the results for R-NL-BSVM are close to that for NL-SVM. This is not surprising, as a pure classification model (e.g., NL-SVM) does not attempt to explain the data but only to maximize classification performance. In this case, the most expensive approach, namely R-NL-BSVM has a runtime in the neighborhood of 10 minutes which is deemed acceptable considering the size of the dataset. Visualizations of the factor loadings, ${\bf A}$, learned by various models are presented in the Supplementary Material.
			\begin{table*}[ht!]
				\caption{AUC (with error bars), accuracy and runtime in seconds for gene expression data.}
				\centering
				\scalebox{0.86}{
					\begin{tabular}{c|ccccc}
						Methods & PFA-L-BSVM  & G-L-BSVM & R-L-BSVM & G-NL-BSVM & R-NL-BSVM \\
						\hline
						TB vs. Others & 0.740$\pm$0.102, 0.683 & 0.766$\pm$0.093, 0.704 & 0.814$\pm$0.052, 0.740  & 0.847$\pm$0.061, 0.778 & {\bf 0.872$\pm$0.025, 0.781} \\
						Active TB vs. Others & 0.802$\pm$0.070, 0.775 & 0.857$\pm$0.050, 0.784 &  0.896$\pm$0.028, 0.832  & 0.921$\pm$0.034, 0.853 & {\bf 0.948$\pm$0.021, 0.880} \\
						Latent TB vs. Others  & 0.849$\pm$0.051, 0.802 & 0.907$\pm$0.037, 0.841 & 0.923$\pm$0.041, 0.868  & 0.934$\pm$0.029, 0.874 & {\bf 0.954$\pm$0.025, 0.889} \\
						HIV(+) vs. HIV(-) & 0.850$\pm$0.056, 0.793 & 0.879$\pm$0.055, 0.844 & 0.900$\pm$0.055, 0.856  & 0.915$\pm$0.041, 0.850 & {\bf 0.959$\pm$0.051, 0.901} \\
						One fold time & 130 & 141 & 180  & 330 & 450 \\
					\end{tabular}}
					\label{Table:TB}
					\vspace{-3mm}
				\end{table*}
				\subsection{Gene expression}\label{sc:tb}
				We applied our model to a newly published tuberculosis study from \citet{Anderson14}, consisting of gene expression intensities for $47323$ genes and 334 subjects (GEO accession series GSE39941). These subjects can be partitioned in three phenotypes: active tuberculosis (TB) (111), latent TB (54) and other infectious diseases (169), and in whether they are positive (107) or negative (227) for HIV. The raw data were preprocessed with background correction, sample-wise scaling and gene filtering. For the analysis we keep the top $4732$ genes with largest intensity profiles. Results for three binary classification tasks using a \emph{one vs. the rest} scheme, the HIV classifier and 10-fold cross-validation are summarized in Table~\ref{Table:TB} (error bars for accuracies omitted due to space limitations). We also present area under the ROC curve (AUC) to account for subset imbalance. As an additional baseline, we included a Poisson Factor Analysis (PFA) model \citep{ZhouAISTATS12} with Bayesian SVMs, as a 2-step procedure. For the Gaussian models we log-transform intensities, and for PFA we round them to the closest integer value (raw intensities become floating point values after background correction and scaling). We can see that our models outperform the others in each of the classification subtasks, and R-NL-BSVM performs the best overall with a reasonable computational cost. It is important to mention that we are not building separate discriminative factor models for each subtask, instead a single factor model jointly learns the four predictors, meaning that all classifiers share the same loadings and factor scores. As a result, our model operates here as a \emph{multi-tasking learning} scheme.
				\begin{figure}[htbp!]
					\centering
					\vspace{-2mm}
					\begin{psfrags}
						\psfrag{weight}[c][c][0.6][0]{Weight}
						\psfrag{factor}[c][c][0.6][0]{Factor}
						\psfrag{active-tb-vs-others}[l][l][0.45][0]{Active TB vs. Others}
						\psfrag{latent-tb-vs-others}[l][l][0.45][0]{Latent TB vs. Others}
						\psfrag{tb-vs-others}[l][l][0.45][0]{TB vs. Others}
						\psfrag{hivppp-vs-hivmmm}[l][l][0.45][0]{HIV(+) vs HIV(-)}
						\includegraphics[scale=0.29]{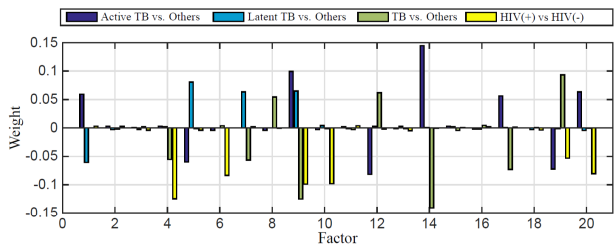}
					\end{psfrags}
					\vspace{-4mm}
					\caption{The learned coefficients $\boldsymbol{\beta}$ for the 4 classifiers based on gene expression data.}
					\label{fg:classifier}
				\end{figure}
				Figure~\ref{fg:classifier} shows the coefficients of the classifier learned using R-L-BSVM. The leading coefficients reveal that certain factors are key to different classes. For instance, factor 14 is specific to TB vs. others, factors 1 and 5 are specific to TB vs. latent TB, and factors 9 and 4 are specific to TB vs. others including HIV(+). We performed a pathway association analysis using DAVID \citep{Huang09Nature} on the top 200 genes from each factor. We found interesting associations. Factor 14: ubiquitin-protein, ligase activity and immunodeficiency. Factor 9: immune response, lymphocite/leukocite/T cell activation and apoptosis. Factor 5: proteasome complex, response to stress, response to antibiotic. Factor 4: ribonucleoprotein, proteasome, ubiquitin-protein, ligase activity. The complete gene lists and the inferred gene networks are provided in the Supplementary Material.
				%
				\subsection{RNA sequencing}
				Finally, we consider a new RNA sequencing (RNAseq) sepsis study \citep{langley2013integrated}. The dataset consists of 133 subjects and 15158 genes (after removing genes with more than 25\% zero entries). Data preprocessing consists of sample-wise rescaling to compensate for differences in library size, log-transform for Gaussian models and rounding for PFA. Subjects are split into three different groups, namely systemic inflammatory response (SIRS) (26), sepsis survivors (SeS) (78) and sepsis complications leading to death (SeD) (29). Three binary comparisons are the main interest of the study: SIRS vs (all) sepsis, SeS vs. SeD and SIRS vs. SeS. Being able to classify this sub-groupings is important for two reasons: ($i$) these tasks are known to be hard classification problems, and ($ii$) a recently published study by \citet{lui14a} showed that approximately 40\% of hospital mortality is sepsis related. Classification results for 10-fold cross-validation including AUC, accuracy and runtime per fold are summarized in Table~\ref{Table:capsod}. Once again our model performs the best. We also tried the nonlinear version of our model but figures were omitted due to very minor improvements in performance.
				\begin{table}[ht!]
					\vspace{-5mm}
					\caption{AUC, accuracy and runtime(s) for RNAseq data.}
					\centering
					\resizebox{1.0\columnwidth}{!}
					{
						\begin{tabular}{c|ccc}
							Methods & PFA-L-BSVM & G-L-BSVM & R-L-BSVM  \\
							\hline
							SIRS vs. Se & 0.70$\pm$0.04, 0.73 & 0.78$\pm$0.02, 0.76 & {\bf 0.86$\pm$0.01, 0.81}  \\
							SeD vs. SeS & 0.76$\pm$0.05, 0.70 & 0.76$\pm$0.01, 0.75 &  {\bf  0.82$\pm$0.02, 0.78}  \\
							SIRS vs. SeS & 0.75$\pm$0.02, 0.71 & 0.87$\pm$0.01, 0.71  & {\bf 0.91$\pm$0.01, 0.87}  \\
							One fold time & 179 & 175 & 226 \\
						\end{tabular}
					}
					\label{Table:capsod}
					\vspace{-4mm}
				\end{table}
				
				\vspace{-5mm}
				\section{Conclusion}
				\vspace{-1mm}
				We have developed a Bayesian discriminative factor model for data that are generally non-Gaussian. This is achieved via the integration of a new max-margin rank likelihood, Bayesian support vector machines, global-local shrinkage priors, and a Dirichlet process mixture model. The proposed model is built on the ranks of the data, opening the door to treat ordinal, continuous and discrete data (\eg count data) within the same framework. Experiments have demonstrated that the proposed factor model achieves better performance than widely used log-transformed-plus-Gaussian models and a Poisson model, on both gene expression and RNA sequencing data. These results highlight the potential of the proposed model in a variety of applications, especially computational biology.
				
				Our rank based models are relatively more computationally expensive than Gaussian models on log-transformed data. However, in applications such as gene expression or sequencing that constitute the real data used in our experiments, runtimes are still significantly lower when compared to the time needed to generate the data. For biological studies, the quality and interpretability of the results are of paramount importance, with speed a secondary issue.
\vspace{-3mm}
\section*{Acknowledgments}
\vspace{-2mm}
The research reported here was funded in part by ARO, DARPA, DOE, NGA and ONR.


\begin{thebibliography}{32}
	\providecommand{\natexlab}[1]{#1}
	\providecommand{\url}[1]{\texttt{#1}}
	\expandafter\ifx\csname urlstyle\endcsname\relax
	\providecommand{\doi}[1]{doi: #1}\else
	\providecommand{\doi}{doi: \begingroup \urlstyle{rm}\Url}\fi
	
	\bibitem[Anderson~{\em et. al.}(2014)]{Anderson14}
	Anderson~{\em et. al.}, S.~T.
	\newblock Diagnosis of childhood tuberculosis and host {RNA} expression in
	{Africa}.
	\newblock \emph{The New England Journal of Medicine}, 370\penalty0
	(18):\penalty0 1712--1723, 2014.
	
	\bibitem[Armagan et~al.(2011)Armagan, Clyde, and Dunson]{Armagan12}
	Armagan, A., Clyde, M., and Dunson, D.~B.
	\newblock Generalized beta mixtures of gaussians.
	\newblock In \emph{NIPS 24}, 2011.
	
	\bibitem[Bhattacharya \& Dunson(2011)Bhattacharya and Dunson]{Dunson11MultiGam}
	Bhattacharya, A. and Dunson, D.~B.
	\newblock Sparse {B}ayesian infinite factor models.
	\newblock \emph{Biometrika}, 98\penalty0 (2):\penalty0 291--306, 2011.
	
	\bibitem[Blei \& Jordan(2005)Blei and Jordan]{Blei05VB}
	Blei, D.~M. and Jordan, M.~I.
	\newblock Variational inference for dirichlet process mixtures.
	\newblock \emph{Bayesian Analysis}, 1:\penalty0 121--144, 2005.
	
	\bibitem[Carvalho et~al.(2010)Carvalho, Polson, and Scott]{Carvalho_horseshoe}
	Carvalho, C.~M., Polson, N.~G., and Scott, J.~G.
	\newblock The horseshoe estimator for sparse signals.
	\newblock \emph{Biometrika}, 97\penalty0 (2):\penalty0 465--480, 2010.
	
	\bibitem[Chib et~al.(1998)Chib, Greenberg, and Winkelmann]{chib1998posterior}
	Chib, S., Greenberg, E., and Winkelmann, R.
	\newblock Posterior simulation and {B}ayes factors in panel count data models.
	\newblock \emph{Journal of Econometrics}, 86:\penalty0 33--54, 1998.
	
	\bibitem[Dillies et~al.(2013)]{dillies13a}
	Dillies, M.-A. et~al.
	\newblock A comprehensive evaluation of normalization methods for illumina
	high-throughput rna sequencing data analysis.
	\newblock \emph{Briefings in bioinformatics}, 14\penalty0 (6):\penalty0
	671--683, 2013.
	
	\bibitem[Fu et~al.(2010)Fu, Robles-Kelly, and Zhou]{Fu10TNN}
	Fu, Z., Robles-Kelly, A., and Zhou, J.
	\newblock Mixing linear {SVM}s for nonlinear classification.
	\newblock \emph{IEEE-TNN}, 2010.
	
	\bibitem[Henao et~al.(2014)Henao, Yuan, and Carin]{Henao2014}
	Henao, R., Yuan, X., and Carin, L.
	\newblock Bayesian nonlinear support vector machines and discriminative factor
	modeling.
	\newblock In \emph{NIPS}, 2014.
	
	\bibitem[Hoff(2007)]{hoff07a}
	Hoff, P.~D.
	\newblock Extending the rank likelihood for semiparametric copula estimation.
	\newblock \emph{The Annals of Applied StatisticsS}, 1\penalty0 (1):\penalty0
	265--283, 2007.
	
	\bibitem[Hoff(2009)]{hoff2009}
	Hoff, P.~D.
	\newblock \emph{A First Course in Bayesian Statistical Methods}.
	\newblock Springer, 2009.
	
	\bibitem[Huang et~al.(2009)Huang, Sherman, and Lempicki]{Huang09Nature}
	Huang, D.~W., Sherman, B.~T., and Lempicki, R.A.
	\newblock Systematic and integrative analysis of large gene lists using david
	bioinformatics resources.
	\newblock \emph{Nature Protoc.}, 1:\penalty0 44--57, 2009.
	
	\bibitem[Ishwaran \& James(2001)Ishwaran and James]{Ishwaran01DP}
	Ishwaran, H. and James, L.~F.
	\newblock Gibbs sampling methods for stick-breaking priors.
	\newblock \emph{JASA}, 96\penalty0 (453):\penalty0 161--173, 2001.
	
	\bibitem[Lehmann \& D'Abrera(2006)Lehmann and D'Abrera]{lehmann06a}
	Lehmann, E.~L. and D'Abrera, H.~J.~M.
	\newblock \emph{Nonparametrics: statistical methods based on ranks}.
	\newblock Springer New York, 2006.
	
	\bibitem[Liu et~al.(2014)Liu, Escobar, and Greene~{\em et al. }]{lui14a}
	Liu, V., Escobar, G~.J., and Greene~{\em et al. }, J.~D.
	\newblock Hospital deaths in patients with sepsis from 2 independent cohorts.
	\newblock \emph{Journal of the American Medical Association}, 2014.
	
	\bibitem[MacKay(1995)]{mackay95a}
	MacKay, D.~J.~C.
	\newblock Probable networks and plausible predictions -- {A} review of
	practical {B}ayesian methods for supervised neural networks.
	\newblock \emph{Network: Computation in Neural Systems}, 6\penalty0
	(3):\penalty0 469--505, 1995.
	
	\bibitem[Mairal et~al.(2008)Mairal, Bach, Ponce, Sapiro, and
	Zisserman]{Mairal08superviseddictionary}
	Mairal, J., Bach, F., Ponce, J., Sapiro, G., and Zisserman, A.
	\newblock Supervised dictionary learning.
	\newblock In \emph{NIPS 21}, 2008.
	
	\bibitem[Monahan \& Boos(1992)Monahan and Boos]{monahan92a}
	Monahan, J.~F. and Boos, D.~D.
	\newblock Proper likelihoods for {B}ayesian analysis.
	\newblock \emph{Biometrika}, 79\penalty0 (2):\penalty0 271--278, 1992.
	
	\bibitem[Murray et~al.(2013)Murray, Dunson, Carin, and
	Lucas]{murray2013bayesian}
	Murray, J.~S., Dunson, D.~B., Carin, L., and Lucas, J.~E.
	\newblock Bayesian {G}aussian copula factor models for mixed data.
	\newblock \emph{JASA}, 108\penalty0 (502):\penalty0 656--665, 2013.
	
	\bibitem[Neal(2000)]{neal00a}
	Neal, R.~M.
	\newblock Markov chain sampling methods for {D}irichlet process mixture models.
	\newblock \emph{Journal of Computational and Graphical Statistics}, 9\penalty0
	(2):\penalty0 249--265, 2000.
	
	\bibitem[Pettitt(1982)]{Pettitt82rank}
	Pettitt, A.~N.
	\newblock Inference for the linear model using a likelihood based on ranks.
	\newblock \emph{JRSS-B}, 44\penalty0 (2):\penalty0 234--243, 1982.
	
	\bibitem[Polson \& Scott(2010)Polson and Scott]{polson10a}
	Polson, N.~G. and Scott, J.~G.
	\newblock Shrink globally, act locally: sparse {B}ayesian regularization and
	prediction.
	\newblock \emph{Bayesian Statistics}, 9:\penalty0 501--538, 2010.
	
	\bibitem[Polson \& Scott(2011{\natexlab{a}})Polson and
	Scott]{polson11Rejoinder}
	Polson, N.~G. and Scott, S.~L.
	\newblock Rejoinder: Data augmentation for support vector machines.
	\newblock \emph{Bayesian Analysis}, 6\penalty0 (1):\penalty0 43--48,
	2011{\natexlab{a}}.
	
	\bibitem[Polson \& Scott(2011{\natexlab{b}})Polson and Scott]{polson11a}
	Polson, N.~G. and Scott, S.~L.
	\newblock Data augmentation for support vector machines.
	\newblock \emph{Bayesian Analysis}, 6\penalty0 (1):\penalty0 1--23,
	2011{\natexlab{b}}.
	
	\bibitem[Quadrianto et~al.(2013)Quadrianto, Sharmanska, Knowles, and
	Ghahramani]{Quadriantos13IBP}
	Quadrianto, N., Sharmanska, V., Knowles, D.~A., and Ghahramani, Z.
	\newblock The supervised {IBP}: Neighbourhood preserving infinite latent
	feature models.
	\newblock \emph{CoRR}, 2013.
	
	\bibitem[Salazar et~al.(2012)Salazar, Cain, Mitroff, and Carin]{Salazar12ICML}
	Salazar, E., Cain, M.~S., Mitroff, S.~R., and Carin, L.
	\newblock Inferring latent structure from mixed real and categorical relational
	data.
	\newblock In \emph{ICML}, 2012.
	
	\bibitem[Scholkopf \& Smola(2001)Scholkopf and Smola]{scholkopf01a}
	Scholkopf, B. and Smola, A.~J.
	\newblock \emph{Learning with kernels: support vector machines, regularization,
		optimization, and beyond}.
	\newblock MIT press, 2001.
	
	\bibitem[Sethuraman(2001)]{Sethuraman01DP}
	Sethuraman, J.
	\newblock A constructive definition of the {D}irichlet prior.
	\newblock \emph{Statistica Sinica}, 4:\penalty0 639--650, 2001.
	
	\bibitem[Shahbaba \& Neal(2009)Shahbaba and Neal]{shahbaba09a}
	Shahbaba, B. and Neal, R.~M.
	\newblock Nonlinear models using {D}irichlet process mixtures.
	\newblock \emph{JMLR}, 10:\penalty0 1829--1850, 2009.
	
	\bibitem[Wipf \& Nagarajan(2008)Wipf and Nagarajan]{Wipf08anew}
	Wipf, D. and Nagarajan, S.
	\newblock A new view of automatic relevance determination.
	\newblock In \emph{NIPS 21}, 2008.
	
	\bibitem[Xu et~al.(2013)Xu, Zhu, and Zhang]{xu2013fast}
	Xu, M., Zhu, J., and Zhang, B.
	\newblock Fast max-margin matrix factorization with data augmentation.
	\newblock In \emph{ICML}, 2013.
	
	\bibitem[Zhou et~al.(2012)Zhou, Hannah, Dunson, and Carin]{ZhouAISTATS12}
	Zhou, M., Hannah, L., Dunson, D., and Carin, L.
	\newblock Beta-negative binomial process and {P}oisson factor analysis.
	\newblock In \emph{AISTATS}, 2012.
	
\end{thebibliography}
\bibliographystyle{icml2015}

\section{Model}
The full Bayesian model is:
\begin{align*}
x_{i,n} &= \ g_i(w_{i,n}), \quad i = 1,\dots,d; ~~ n = 1,\dots,N; \\
L_i(w_{i,n}|\w_{i\bs n})&= \int \DN(w_{i,n}-w_{i,n}^u;-\epsilon-\lambda_{i,n}^u,\lambda_{i,n}^u)\\ & \times\DN(w_{i,n}^l-w_{i,n};-\epsilon-\lambda_{i,n}^l,\lambda_{i,n}^l)d\lambda_{i,n}^u d\lambda_{i,n}^l \,, \\
L_n(y_n|\bbeta,\z_n) &= \int_0^{\infty} \frac{1}{\sqrt{2\pi \lambda^c_n}} \\
&\times \exp\left( -\frac{1}{2} \frac{(1-y_n\bbeta\ts\z_n + \lambda^c_n)^2}{\lambda^c_n}\right) d\lambda^c_n \,, \\
\z_n \ &\sim \ \DN(\bmu_{t(n)},\psi_{t(n)}\inv\I_K) \,, \\
\bmu_{t}  &\sim({\bf 0}, \I_K), \\
\psi_t &\sim {\rm Ga}(\psi_s,\psi_r), \\
\quad t(n) &\sim {\rm Mult}(1; q_1,\dots, q_T), \\
q_t &= \nu_t\prod_{l=1}^{t-1}(1-\nu_l) \,, \\
\nu_t &\sim {\rm Beta}(1,\alpha), \quad \alpha \sim{\rm Ga}(\alpha_s,\alpha_r), \\
a_{i,k}  \ &\sim  \ \DN(0, \xi_{i,k}) \,,  \quad \xi_{i,k} \ \sim  \ \Ga(r_a, \eta_{i,k}) \,,\\
\eta_{i,k} \ &\sim  \ \Ga(s_a,\Phi^{(a)}_k ) \,, \\
\Phi^{(a)}_k &\sim \Ga(1/2,\tilde{\Phi}^{(a)}), ~\quad \tilde{\Phi}^{(a)} \sim {\rm Ga}(1/2,1),\\
\beta_{k} \ &\sim  \ \DN(0, b_k) \,,  \qquad b_k \ \sim  \ \Ga(r_{\beta}, e_k) \,, \\
e_k \ &\sim  \ \Ga(s_{\beta},\Phi^{(\beta)} ) \,,\\
\Phi^{(\beta)} &\sim \Ga(1/2,\tilde{\Phi}^{(\beta)}),~\quad \tilde{\Phi}^{(\beta)} \sim {\rm Ga}(1/2,1).
\end{align*}

\section{MCMC inference}
For convenience, we denote
\begin{eqnarray*}
	\lambda_{i,n} &=&(\lambda_{i,n}^l)\inv+(\lambda_{i,n}^u)\inv,\\
	\Delta^{(k)}_{i,n}  &=& \left(\frac{w_{i,n}^l + \epsilon - w_{i,n}}{\lambda^l_{i,n}} - \frac{w_{i,n} + \epsilon - w_{i,n}^u}{\lambda^u_{i,n}}\right) \\
	&&+ a_{i,k} z_{k,n}\lambda_{i,n},
\end{eqnarray*}
\begin{eqnarray*}
	(\bbeta\ts\z_n)_{\bs k} & =&\bbeta\ts\z_n-\beta_kz_{k,n},\\
	\Gamma_{k,n} &=& \frac{y_n\beta_k [1+\lambda_n^c - y_n (\bbeta\ts\z_n)_{\bs k}] }{\lambda^c_n}.
\end{eqnarray*}
In the following conditional posterior-distributions, \quotes{$\cdot$} refers to the conditioning parameters of the distributions, ${\rm IG}(a,b)$ denotes the inverse Gaussian distribution, ${\rm Ga}(a,b)$ the gamma distribution, and ${\rm GIG}(a,b,p)$ the generalized inverse Gaussian distribution. 

In the {\em linear} case, when the Dirichlet process mixture (DPM) model is not used, the conditional posterior distributions are:
\begin{enumerate}
	\item ${\bf A}$:
	\begin{align*}
	p(a_{i,k}| \ \cdot)  &=  \ \DN(\mu_{a_{i,k}}, \sigma^2_{a_{i,k}}) \,,  \\
	\sigma^{-2}_{a_{i,k}}& = \xi_{i,k}^{-1} + \sum_{n=1}^N z^2_{k,n} \lambda_{i,n}\inv \,,\\
	\mu_{a_{i,k}} &= \sigma^2_{a_{i,k}} \sum_{n=1}^N z_{k,n} \Delta^{(k)}_{i,n} \,, \\
	p(\xi_{i,k}|\cdot)&= {\rm GIG}(2\eta_{i,k}, a_{i,k}^2, r_a - 0.5) \,, \\ 
	p(\eta_{i,k}|\cdot)&= {\rm Ga}(r_a + s_a, \xi_{i,k} + \Phi^{(a)}_k) \,, \\
	p(\Phi^{(a)}_k|\cdot) &= {\rm Ga}\left(\frac{1}{2}+ s_a d, \tilde{\Phi}^{(a)} + \frac{1}{2} \sum_{i} \eta_{i,k}\right), \\
	p(\tilde{\Phi}^{(a)}|\cdot) &= {\rm Ga}\left(1, \sum_k\Phi^{(a)}_k +1\right) \,.
	\end{align*}
	
	\item ${\bf Z}$:
	\begin{align*}
	p(z_{k,n}| \ \cdot) &=  \ \DN(\mu_{z_{k,n}}, \sigma^2_{z_{k,n}}) \,, \\
	\sigma^{-2}_{z_{k,n}} &= 1 + \sum_{i=1}^d  a^2_{i,k}\lambda_{i,n}\inv + \frac{\beta^2_k}{\lambda^c_n} \,, \\
	\mu_{z_{k,n}} &= \sigma^2_{z_{k,n}}\left(\sum_{i=1}^d a_{i,k} \Delta^{(k)}_{i,n} + \Gamma_{k,n}\right) \,.
	\end{align*}
	
	\item ${{\bf \Lambda}^l, {\bf \Lambda}^u}$, $\boldsymbol{\lambda}^c$:
	\begin{align*}
	p\left((\lambda_{i,n}^l)^{-1}|\cdot\right) &= {\rm IG}\left(|w_{i,n}^l + \epsilon - w_{i,n}|^{-1},1\right)\,,\\
	p\left((\lambda_{i,n}^u)^{-1}|\cdot\right) &= {\rm IG}\left(|w_{i,n} + \epsilon - w_{i,n}^u|^{-1},1\right)\,,\\
	p\left((\lambda_{n}^c)^{-1}|\cdot\right) &= {\rm IG}\left(|1-y_n \bbeta\ts\z_n|^{-1},1\right)\,.
	\end{align*}
	
	\item $\boldsymbol{\beta}$:
	\begin{align*}
	p(\beta_k|\cdot)&= {\cal N}(\mu_{\beta_k},\sigma^2_{\beta_k}) \,, 
	\quad \sigma^{-2}_{\beta_k} = b_k^{-1} + \sum_{n=1}^N \frac{z^2_{k,n}}{\lambda_n^c} \,, \\ 
	\quad \mu_{\beta_k} &= \sigma^2_{\beta_k}\sum_{n=1}^N \frac{y_n z_{k,n} \left[1+\lambda_n^c - y_n (\bbeta\ts\z_n)_{\bs k}\right]}{\lambda_n^c} \,, \\
	p(b_k|\cdot)&= {\rm GIG}(2 e_k, \beta_k^2, r_{\beta} - 0.5), \\
	p(e_k|\cdot)&= {\rm Ga}(r_{\beta} + s_{\beta}, b_k + \Phi^{(\beta)}_k) \,, \\
	p(\Phi^{(\beta)}|\cdot) &= {\rm Ga}\left(\frac{1}{2}+ s_{\beta}K, \tilde{\Phi}^{(\beta)} + \frac{1}{2} \sum_k e_k\right) \,, \\
	p(\tilde{\Phi}^{(\beta)}|\cdot) &= {\rm Ga}\left(1, \Phi^{(\beta)} +1\right) \,.
	\end{align*}
\end{enumerate}

In the {\em nonlinear} case, when the DPM is used:
\begin{enumerate}
	\item $t(n)$ (mixture component index for $n$-th observation):
	\begin{align*}
	p(t(n) = t | \cdot) \propto q_t {\cal N}({\bf z}_n; \bmu_{t},\psi_{t}\inv\I_K) \,.
	\end{align*}
	
	
	\item DPM parameters:
	\begin{align*}
	p(\mu_{t,k}|\cdot) &= {\cal N}(\mu_{\mu_{t,k}}, \sigma^2_{\mu_{t,k}}) \,, \quad
	\sigma^{-2}_{\mu_{t,k}}  = 1 +  \sum_{n:t(n) = t} \psi_t \,,\\
	\mu_{\mu_{t,k}} &= \sigma^{2}_{\mu_{t,k}} \psi_t \sum_{n:t(n) = t} z_{k,n} \,, \\
	p(\psi_t|\cdot) &= {\rm Ga}\left(\psi_s + 0.5 K, ~~\psi_r + 0.5 \sum_k \mu_{t,k}^2\right) \,, \\
	p(\nu_t|\cdot) & = {\rm Beta}\left(1 + \sum_{n:t(n) = t} 1, ~~\alpha + \sum_{n:t(n)>t} 1\right) \,, \\
	p(\alpha|\cdot) &= {\rm Ga}\left(\alpha_s + T-1, ~~\alpha_r -\sum_{t=1}^{T-1}\log (1-\nu_t) \right) \,.
	\end{align*}
	
\end{enumerate}
In this case, $\boldsymbol{\beta}$ in 2) and 4) should be replaced by $\boldsymbol{\beta}^{(t)}$, for $t=1,\dots, T$, and the summation over $n$ in 4) will only account for $\{n:t(n)=t\}$.

\section{VB inference}
Since the model is fully local conjugate, the VB update equations can be obtained using the moments of the above conditional posterior distributions. Here we present the moments for the model without DPM, and for the VB inference of the DP mixture model, please refer to~\cite{Blei05VB}. In the following expressions, $\langle\cdot\rangle$ denotes expectation, ${\cal K}_p(\cdot)$ is the modified Bessel function of the second kind, $\langle w_{i,n}\rangle = \langle {\bf a}\ts_i \rangle \langle {\bf z}_n \rangle$, $\langle w^l_{i,n}\rangle = \langle {\bf a}\ts_i \rangle \langle {\bf z}^l_n \rangle$ and $\langle w^u_{i,n}\rangle = \langle {\bf a}\ts_i \rangle \langle {\bf z}^u_n \rangle$.
\begin{enumerate}
	\item ${\bf A}$:
	\begin{align*}
	\langle a_{i,k}\rangle & = \langle \sigma^2_{a_{i,k}} \rangle \sum_{n=1}^N \langle z_{k,n} \rangle \langle \Delta_{i,n}^{(k)} \rangle,\\
	\langle \sigma^{2}_{a_{i,k}} \rangle &= \left(\langle \xi_{i,k}^{-1}\rangle + \sum_{n=1}^N \langle z^2_{k,n}\rangle  \langle \lambda_{i,n}\inv \rangle\right)^{-1}, \\
	\langle a^2_{i,k}\rangle & = \langle a_{i,k}\rangle^2 + \langle \sigma^{2}_{a_{i,k}} \rangle, \\
	\langle \Delta_{i,n}^{(k)} \rangle &= \langle (\lambda^l_{i,n})^{-1}\rangle \left(\langle w_{i,n}^l\rangle + \epsilon - \langle w_{i,n}\rangle\right) \\
	&\quad- (\langle\lambda^u_{i,n})^{-1}\rangle \left(\langle w_{i,n}\rangle + \epsilon - \langle w_{i,n}^u\rangle\right)\\
	&\quad + \langle a_{i,k}\rangle \langle z_{k,n}\rangle \left[\langle(\lambda^l_{i,n})^{-1}\rangle + \langle (\lambda^u_{i,n})^{-1}\rangle\right],\\
	\langle \xi_{i,k} \rangle & = \frac{\sqrt{\langle a^2_{i,k} \rangle}{\cal K}_{r_a + 0.5} \left(\sqrt{2\langle \eta_{i,k}\rangle  \langle a^2_{i,k}\rangle }\right)}{\sqrt{2\eta_{i,k}} {\cal K}_{r_a - 0.5} \left(\sqrt{2\langle\eta_{i,k}\rangle \langle a^2_{i,k}\rangle}\right)},
	\\
	\langle \xi^{-1}_{i,k} \rangle  &= \frac{\sqrt{2\langle \eta_{i,k}\rangle }{\cal K}_{r_a - 0.5} \left(\sqrt{2\langle\eta_{i,k}\rangle \langle a^2_{i,k}\rangle }\right)}{\sqrt{\langle a^2_{i,k} \rangle}{\cal K}_{r_a - 1.5} \left(\sqrt{2 \langle \eta_{i,k}\rangle  \rangle a^2_{i,k}\rangle}\right)} , \\
	\langle \eta_{i,k} \rangle &= \frac{r_a + s_a}{\langle \xi_{i,k}\rangle + \langle\Phi_k^{(a)} \rangle},\\
	~~\langle \Phi_k^{(a)} \rangle  &= \frac{0.5 + d s_a}{\langle \tilde{\Phi}^{(a)}\rangle + 0.5 \sum_i \langle \eta_{i,k} \rangle },\\ 
	\langle \tilde{\Phi}^{(a)}\rangle &= \frac{1}{1+\sum_k\langle \Phi_k^{(a)} \rangle }.
	\end{align*}
	
	\item ${\bf Z}$:
	\begin{align*}
	\langle z_{k,n}\rangle &= \langle \sigma^2_{z_{k,n}}\rangle \left(\sum_{i=1}^d \langle a_{i,k}\rangle \langle \Delta^{(k)}_{i,n}\rangle + \langle \Gamma_{k,n}\rangle\right), 
	\\
	\langle \Gamma_{k,n} \rangle &= \langle(\lambda^c_n)^{-1}\rangle \left\{y_n \langle\beta_k\rangle  [\langle(\lambda^c_n)^{-1}\rangle +1 \right.\\
	&\left.- \langle(\lambda^c_n)^{-1}\rangle y_n (\langle \bbeta\ts \rangle \langle\z_n\rangle )_{\bs k}] \right\},\\
	\langle\sigma^{2}_{z_{k,n}}\rangle  &= \left(1 + \sum_{i=1}^d  \langle a^2_{i,k}\rangle \langle\lambda_{i,n}\inv\rangle + \langle\beta^2_k \rangle\langle(\lambda^c_n)^{-1} \rangle \right)^{-1}, \\ 
	\langle z^2_{k,n}\rangle &= \langle z_{k,n}\rangle^2 + \langle\sigma^{2}_{z_{k,n}}\rangle.
	\end{align*}
	
	\item ${{\bf \Lambda}^l, {\bf \Lambda}^u}$, $\boldsymbol{\lambda}^c$:
	\begin{align*}
	\langle (\lambda_{i,n}^l)^{-1}\rangle &= \left|\langle w_{i,n}^l \rangle + \epsilon - \langle w_{i,n}\rangle \right|^{-1},\\ 
	\langle(\lambda_{i,n}^u)^{-1}\rangle &= \left|\langle w_{i,n} \rangle + \epsilon - \langle w_{i,n}^u\rangle\right|^{-1},\\
	\langle(\lambda_{n}^c)^{-1}\rangle  &= \left|1-y_n \langle \bbeta\ts\rangle \langle \z_n \rangle\right|^{-1}.
	\end{align*}
	
	\item $\boldsymbol{\beta}$:
	\begin{align*}
	\langle {\beta_k} \rangle &= \langle \sigma^2_{\beta_k} \rangle \sum_{n=1}^N \left\{y_n \langle z_{k,n}\rangle [\langle (\lambda_n^c)^{-1} \rangle+1  \right.\\
	&\quad\left.- \langle (\lambda_n^c)^{-1} \rangle y_n (\langle \bbeta\ts \rangle  \langle \z_n \rangle )_{\bs k}]\right\},\\
	\langle \sigma^{2}_{\beta_k}\rangle & = \langle b_k^{-1}\rangle + \sum_{n=1}^N \langle z^2_{k,n}\rangle \langle(\lambda_n^c)^{-1}\rangle, \\
	\langle \beta_k^2 \rangle &= \langle {\beta_k} \rangle^2 +  \langle \sigma^{2}_{\beta_k}\rangle,\\
	\langle b_k \rangle & = \frac{\sqrt{\langle \beta^2_k \rangle}{\cal K}_{r_{\beta}+0.5} \left(\sqrt{2\langle e_k\rangle  \langle \beta^2_{k}\rangle }\right)}{\sqrt{2e_k} {\cal K}_{r_{\beta}-0.5} \left(\sqrt{2\langle e_k\rangle \langle \beta_k^2\rangle}\right)},\\
	\langle b^{-1}_{k} \rangle  &= \frac{\sqrt{2e_k} {\cal K}_{r_{\beta}-0.5} \left(\sqrt{2\langle e_k\rangle \langle \beta_k^2\rangle}\right)}{\sqrt{\langle \beta^2_k \rangle} {\cal K}_{r_{\beta}-1.5} \left(\sqrt{2\langle e_k\rangle \langle \beta_k^2\rangle}\right)},\\
	\langle e_k \rangle &= \frac{r_{\beta} + s_{\beta}} {\langle b_k \rangle + \langle \Phi^{(\beta)}_k\rangle},\\
	\langle \Phi^{(\beta)}\rangle &= \frac{0.5 + 0.5 s_{\beta}}{\langle \tilde{\Phi}^{(\beta)} \rangle + 0.5 \sum_k \langle e_k \rangle },
	\\
	\langle \tilde{\Phi}^{(\beta)} \rangle  &= \frac{1}{\langle \Phi^{(\beta)}\rangle +1}.
	\end{align*}
\end{enumerate}

\section{Inferred Factor Loadings on the Handwritten Digits}
\begin{figure*}[p] 
	\centering
	\includegraphics[scale=0.6]{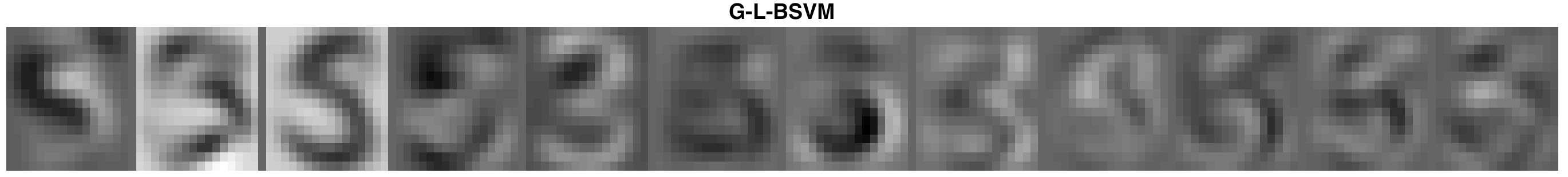}\\
	\includegraphics[scale=0.6]{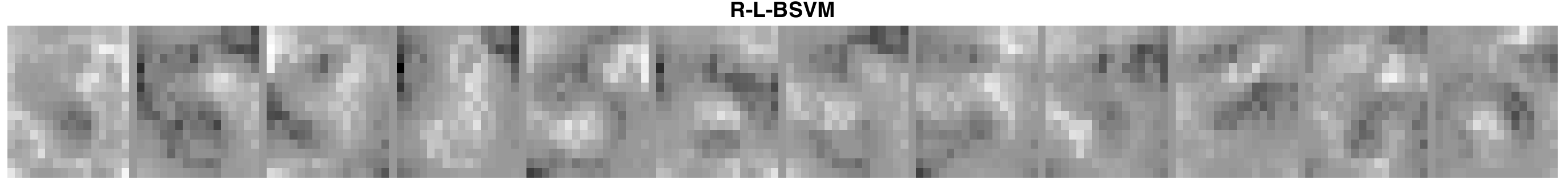}\\
	\includegraphics[scale=0.6]{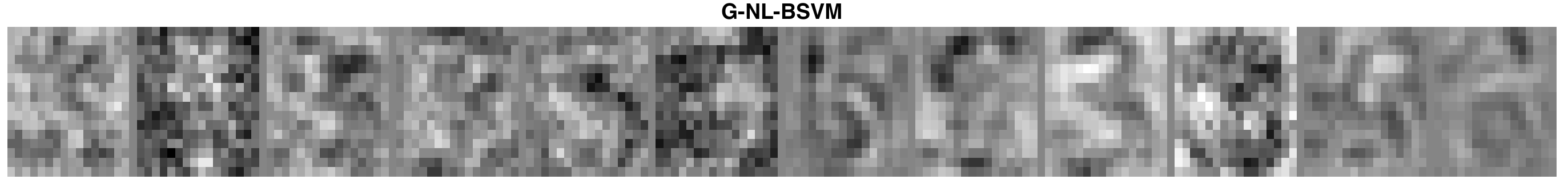}\\
	\includegraphics[scale=0.6]{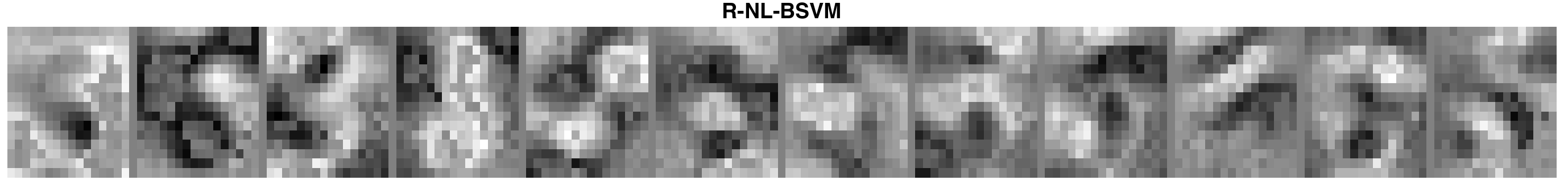}
	\caption{Inferred factor loading matrix $\bf A$ from USPS 3 vs. 5. The first 12 columns are reshaped and plotted.}
	\label{fg:loadings_MNIST}
	\vspace{20mm}
	\includegraphics[scale=0.65]{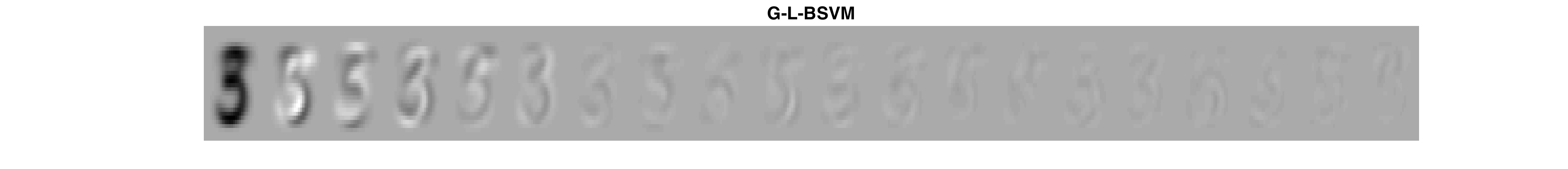}\\
	\includegraphics[scale=0.65]{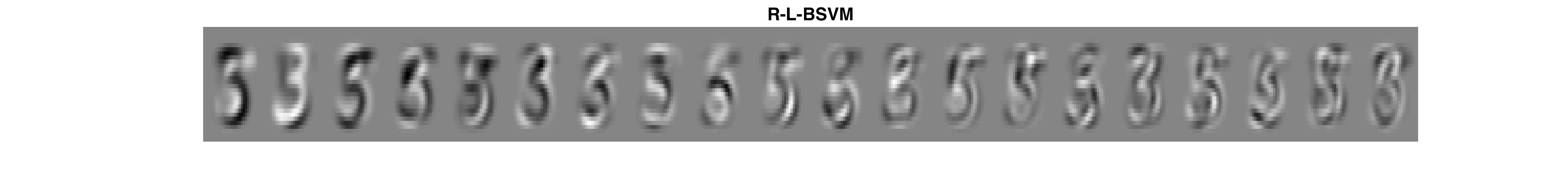}\\
	\includegraphics[scale=0.65]{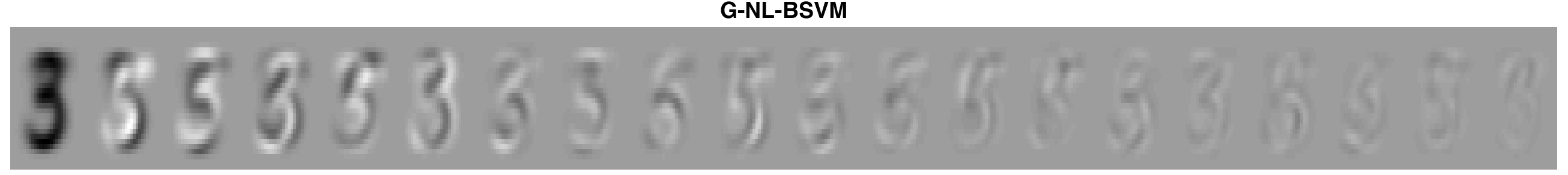}\\
	\includegraphics[scale=0.65]{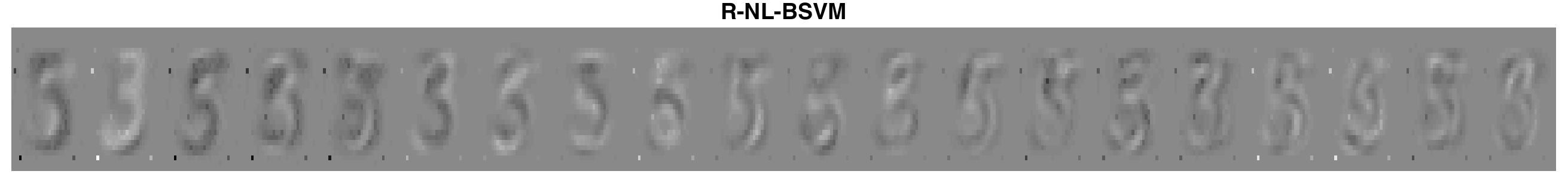}
	\caption{Inferred factor loading matrix $\bf A$ from MNIST 3 vs. 5. The 20 columns are reshaped and plotted.}
	\label{fg:loadings_USPS}
\end{figure*}

We plotted the factor loadings ${\bf A}$ learned from USPS and MNIST datasets in Figures~\ref{fg:loadings_USPS} and \ref{fg:loadings_MNIST}, respectively.
Four models, G-L-BSVM, R-L-BSVM, G-NL-BSVM and R-NL-BSVM are used as examples.
It can be be seen that the Gaussian model is trying to learn the dictionaries to reconstruct images while the rank model is trying to learning the differences (focusing on the edges). 

\begin{figure*}[p] 
	\centering
	\includegraphics[scale=0.29]{coefficent_class1.pdf}
	\caption{\small{The learned classifier coefficients $\boldsymbol{\beta}$ of the 4 classifiers for the gene expression data.}}
	\label{fg:classifier1}
	\vspace{20mm}
	\includegraphics[scale=0.7]{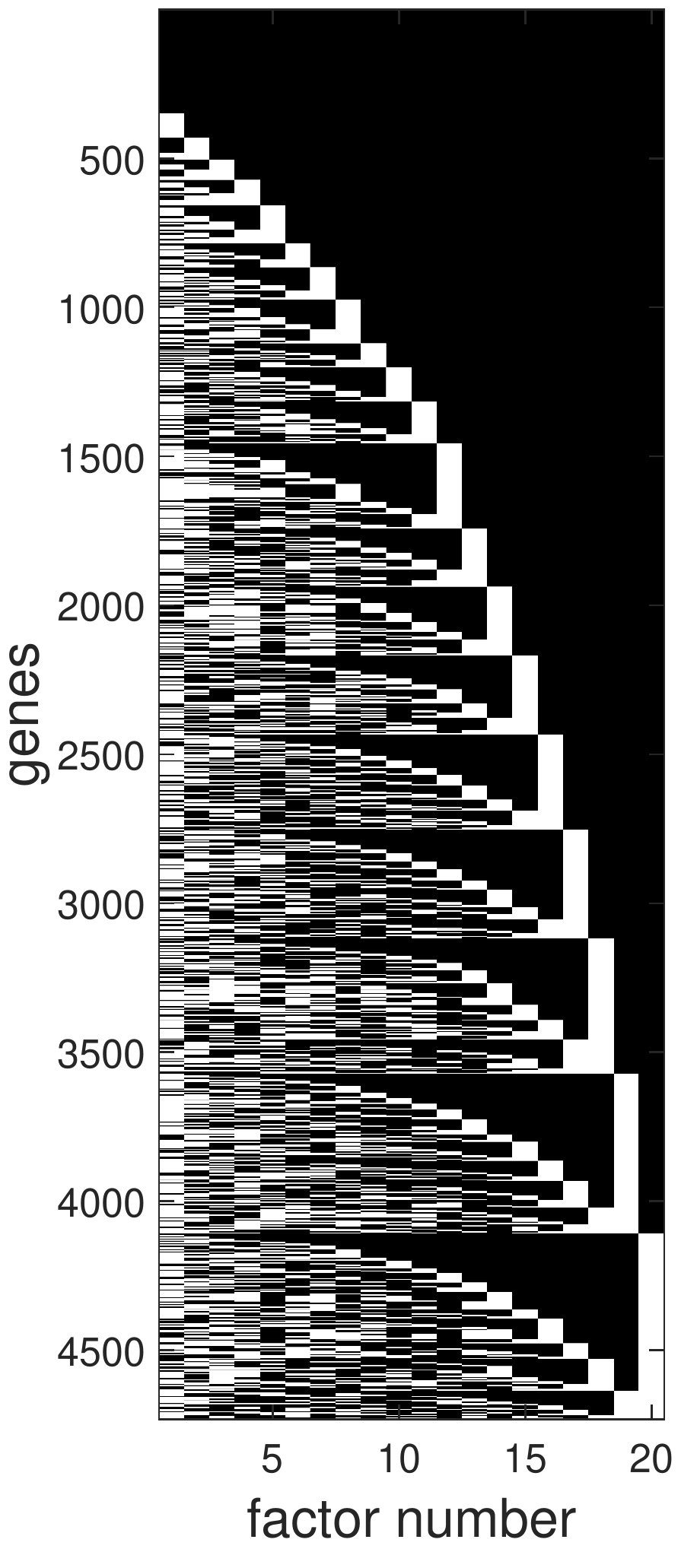}
	\caption{\small{The learned gene network inferred from the factor loading matrix $\bf A$.}}
	\label{fg:loadings}
\end{figure*}

\section{Results on Gene Expression Data}
We show the results of our model for gene expression data. $K=20$ factors are used and here we only show the results generated by the proposed max-margin rank model without DP, {\em i.e.}, using linear Bayesian SVM as the classifier. Figure~\ref{fg:classifier1} shows the coefficients $\boldsymbol{\beta}$ of the learned classifiers and Figure~\ref{fg:loadings} the inferred gene network from the learned factor loading matrix $\bf A$.

%
\end{document}